\documentclass[letterpaper]{article} 
\usepackage{aaai25}  
\usepackage{times}  
\usepackage{helvet}  
\usepackage{courier}  
\usepackage[hyphens]{url}  
\usepackage{graphicx} 
\usepackage{natbib}  
\usepackage{caption} 
\usepackage{amsfonts}
\usepackage{amssymb}
\usepackage{xspace}
\usepackage{booktabs}
\usepackage[ruled,vlined]{algorithm2e}
\usepackage{amsmath,bm}
\usepackage{dsfont}
\usepackage{multirow}
\usepackage [autostyle, english = american]{csquotes}
\usepackage{xcolor}
\usepackage{subcaption} 
\MakeOuterQuote{"}
\usepackage{ntheorem,lipsum} 
\theorembodyfont{\upshape} 

\newcommand{\mname}{\texttt{PatentAgent}\xspace}
\newcommand{\qa}{\emph{PA-QA}\xspace}
\newcommand{\ocsr}{\emph{PA-Img2Mol}\xspace}
\newcommand{\core}{\emph{PA-CoreId}\xspace}

\DeclareCaptionStyle{ruled}{labelfont=normalfont,labelsep=colon,strut=off} 
\usepackage[shortlabels]{enumitem}
\title{\mname: Intelligent Agent for Automated Pharmaceutical Patent Analysis}
\author {
    Xin Wang\textsuperscript{\rm 1*},
    Yifan Zhang\textsuperscript{\rm 1}\footnote{Equal Contribution},
    Xiaojing Zhang\textsuperscript{\rm 1},
    Longhui Yu\textsuperscript{\rm 1},
    Xinna Lin\textsuperscript{\rm 1},
    Jindong Jiang\textsuperscript{\rm 1},
    Bin Ma\textsuperscript{\rm 1}
    Kaicheng Yu\textsuperscript{\rm 1}
}
\affiliations {
    \textsuperscript{\rm 1}Westlake University\\
    buhangyunfei@gmail.com, yifanzhang2024@u.northwestern.edu, dexianz07@gmail.com, longhuiyu98@gmail.com, linxinna@westlake.edu.cn, jiangjinkekao@gmail.com, binma@stu.xjtu.edu.cn, kyu@westlake.edu.cn
}

\begin{document}

\maketitle

\begin{abstract}

Pharmaceutical patents play a vital role in biochemical industries, especially in drug discovery, providing researchers with unique early access to data, experimental results, and research insights. With the advancement of machine learning, patent analysis has evolved from manual labor to tasks assisted by automatic tools. However, there still lacks an unified agent that assists every aspect of patent analysis, from patent reading to core chemical identification. Leveraging the capabilities of Large Language Models (LLMs) to understand requests and follow instructions, we introduce the \textbf{first} intelligent agent in this domain, \mname, poised to advance and potentially revolutionize the landscape of pharmaceutical research. \mname comprises three key end-to-end modules --- \qa, \ocsr, and \core --- that respectively perform (1) patent question-answering, (2) image-to-molecular-structure conversion, and (3) core chemical structure identification, addressing the essential needs of scientists and practitioners in pharmaceutical patent analysis. Each module of \mname demonstrates significant effectiveness with the updated algorithm and the synergistic design of \mname framework. \ocsr outperform existing methods across CLEF, JPO, UOB, and USPTO patent benchmarks with an accuracy gain between 2.46\% and 8.37\% while \core realizes accuracy improvement ranging from 7.15\% to 7.62\% on PatentNetML benchmark. Our code and dataset will be publicly available.

\end{abstract}

\section{Introduction}

Patents provide a unique wealth of information for the early stages of chemical research and development, especially in freedom-to-operate analysis~\cite{mucke2023conduct}, prior-art search~\cite{setchi2021artificial}, and landscape analysis~\cite{ohms2021current}. They play a critical role in lead and target discovery~\cite{zdrazil2024chembl,senger2017assessment}, offering earlier access to innovative insights and biochemical data than publishable journals~\cite{southan2013tracking}. Given the competitive nature of the drug development business~\cite{Bregonje2005PatentsAU}, patents are a vital source of information. Aiming for the wealth of information provided by patents, scientists continuously work to extract relevant contents in patents and analyze them. However, \textit{scientists struggle with extracting and organizing information due to the lack of an unified agent.} Specifically, existing approaches have the following shortcomings:

\textbf{Manual approaches require excessive human effort to extract information.} Traditional methods before the existence of computational technologies, such as manual review and keyword search~\cite{hu2018patent, andres2010patents}, are often viewed as the golden standards in patent analysis. However, these methods require scientists to spend considerable time and effort extracting information. These methods rely heavily on human expertise to interpret complex chemical information, making them expensive and time-consuming.  
\begin{figure*}
    \centering
    \includegraphics[width=\textwidth]{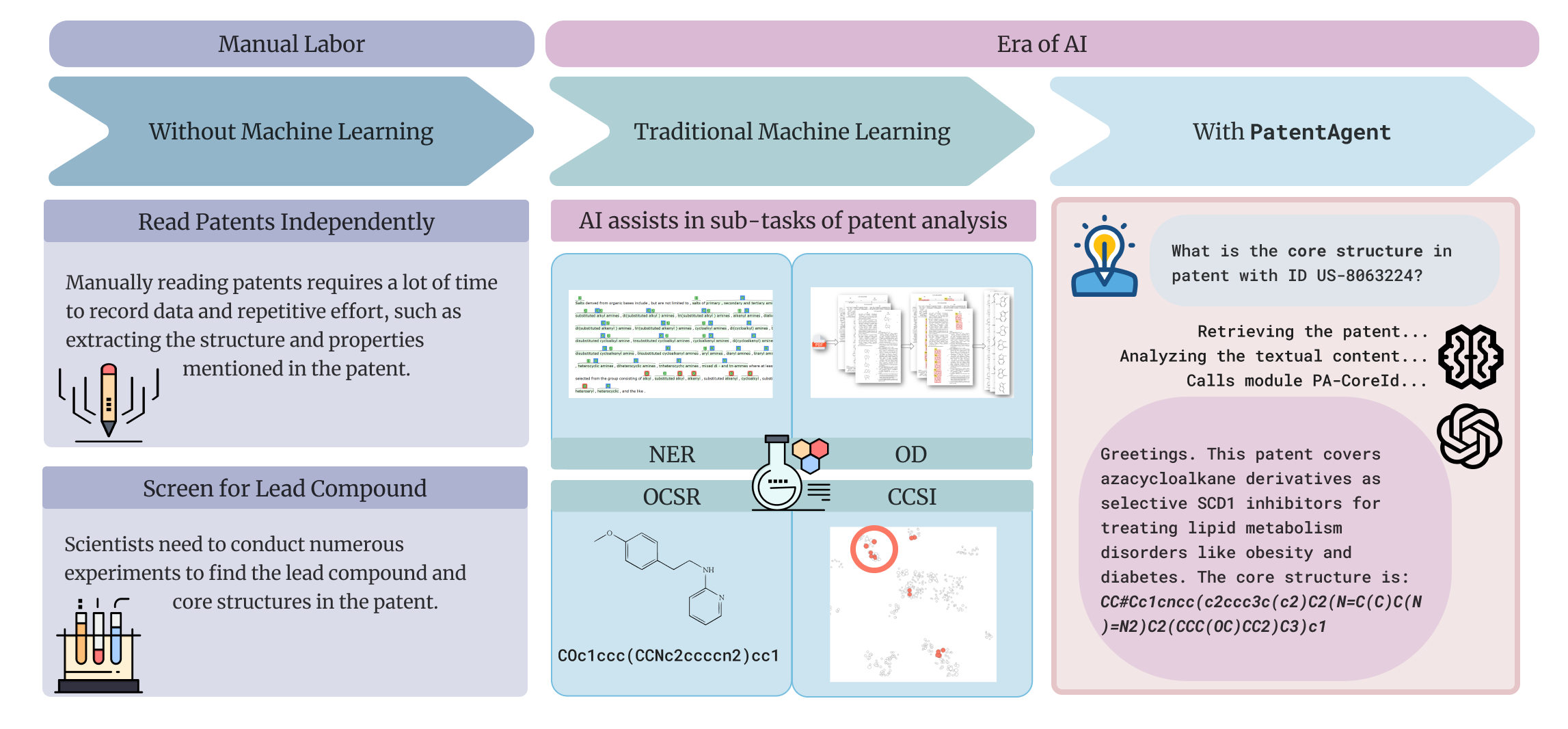}
    \caption{Comparison of patent analysis between manual, traditional machine learning, and \mname. The sub-tasks listed in the \textit{Traditional Machine Learning} phase of patent analysis are: Named Entity Recognition (NER), Object Detection (OD), Optical Chemical Structure Recognition (OCSR), and Core Chemical Structure Identification (CCSI).}
    \label{fig:teaser}
\end{figure*}

\textbf{Current computational tools are highly specialized and lack a holistic solution.} For instance, methods like text mining~\cite{GADIYA2023100069, liu2011development, yun2020automated} and chemical structure exploration~\cite{morin2024patcid, zdrazil2024chembl} operate independently, with little integration or standardized criteria for data formats, algorithms, or evaluation metrics. This lack of common standards makes it difficult to coordinate efforts across multiple modules, a necessity in comprehensive patent analysis. As a result, researchers, particularly those without a background in computer science, such as medical researchers, struggle to effectively utilize these tools. The need for coding skills and the lack of seamless integration create significant barriers, limiting the accessibility and efficiency of these computational methods.

\textbf{Existing tools still struggle to identify the core compound successfully.} Identifying the core compound structure among hundreds or even thousands of chemicals~\cite{habibi2016recognizing} in a pharmaceutical patent is one of the major tasks for scientists and practitioners~\cite{akhondi2019automatic}. Given the complexity of the task, however, current tools struggle at an accuracy level of random guesses~\cite{tyrchan2012exploiting, falaguera2021identification, zhu2024patentnetml}, including the most recent one. 

Those shortcomings make it challenging for AI to assist scientists in analyzing pharmaceutical patents. A unified agent with the capability to precisely interpret and execute scientific requests is essential. To bridge this gap, we aim to create an integrated system that can fully leverage the potential of computational methods to streamline patent analysis~\cite{morin2024patcid}. Fortunately, Large Language Models(LLMs) have demonstrated their exceptional ability to understand human instructions~\cite{zeng2023evaluating} and effectively utilize tools~\cite{ruan2023tptu}, making them well-suited to assist scientists in conducting patent analysis. With the integration of LLMs, \mname intelligently understands human intentions in natural language even when they are multi-fold and involves several patents. 

Besides the LLM orchestrator, \mname consists of three main
modules, \qa, \ocsr, and \core:
\begin{enumerate}[leftmargin=*,noitemsep, topsep=0pt]
    \item \qa: A question-answering chatbot that faithfully responds users' inquiries about patents.
    \item \ocsr: An ensemble of deep learning models and transformer-based models that uses Visual Language Model (VLM) as evaluator. It achieves accuracy gain between 8.37\% and 2.46\% over the best performing models on widely recognized benchmarks of  Optical Chemical Structure Recognition (OCSR), including CLEF, JPO, and UOB.
    \item \core: A machine learning classifier to identify the core chemical structure amongst various chemicals. It realizes better performance on the most updated and accepted PatentNetML dataset for Core Chemical Structure Identification (CCSI).
\end{enumerate}

The success of \mname and its modules across multiple datasets highlights robustness and its ability to consistently outperform existing methods, underscoring its transformative potential. Specifically, \mname not only improves accuracy in critical tasks in chemical structure analysis but also significantly reduces time and effort required to analyze complex pharmaceutical patents. Such effectiveness is further demonstrated in a case study in Section~\ref{sec:case}, where \mname simplifies the patent analysis process, enabling researchers to gain actionable insights with unprecedented pace and reliability.

\section{Method}
\label{sec:method}

\begin{figure*}[t]
\includegraphics[width=\textwidth]{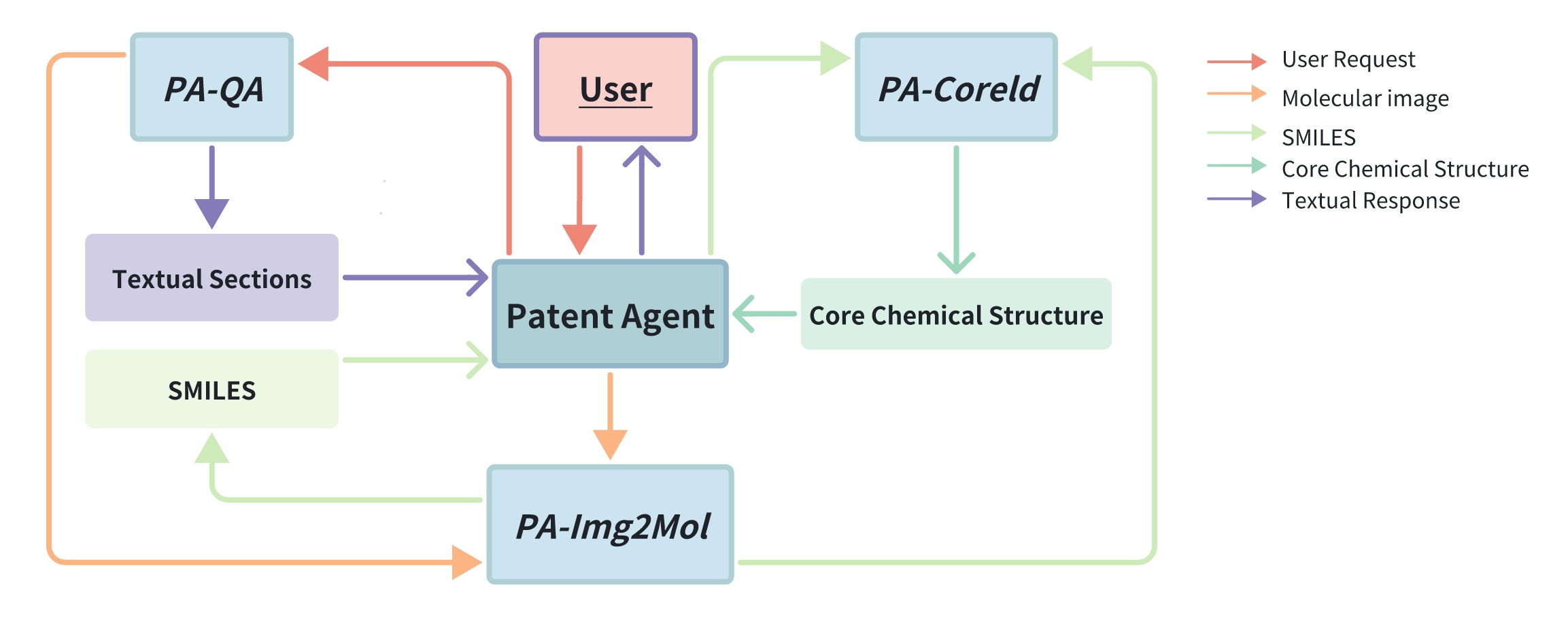}
\caption{Workflow Illustration for \mname. \mname consists of three major modules, namely \qa, \ocsr, and \core. \qa processes user requests and outputs patent segmentation (texts or images), \ocsr processes molecular images and output SMILES, and \core processes SMILES and outputs the core chemical structure. Refer to the color coding legends for better understanding.}
\label{fig:workflow} 
\end{figure*}

This section presents our \mname framework (Fig~\ref{fig:workflow}). Our LLM orchestrator is introduced in Section~\ref{sec:llm}, and the three succeeding subsections discuss three major modules used in \mname. Section~\ref{sec:qa} explains the question answering chatbot \qa that to process natural language queries and generate accurate responses based on the knowledge base of the patent file. Section~\ref{sec:ocsr} describes the module \ocsr and its updated algorithm in converting images of chemical structures to molecular expression with Simplified Molecular Input Line Entry System (SMILES)~\cite{weininger1988smiles} . Section~\ref{sec:core} introduces module \core that identifies the common chemical scaffold across a series of diverse chemical compounds.

\subsection{LLM Orchestrator}
\label{sec:llm}

The LLM orchestrator serves as the central component that connects and coordinates the three major modules: \qa, \ocsr, and \core. Upon receiving a user query, the LLM orchestrator interprets the input and determines whether the query involves retrieving textual information, converting images to molecular structures, or identifying core chemical structures through contextual understanding. It then processes the queries to the requested format and routes the processed queries to corresponding module, aggregates the results and returns a comprehensive response to the user. 

In more complicated cases that span multiple aspects of patent analysis, such as identifying the core chemical structure directly from a patent file, the LLM orchestrator sequences tasks accordingly. It first retrieves chemical images in the pdf files and convert the chemical images to Simplified Molecular Input Line Entry System (SMILES)~\cite{weininger1988smiles} with \ocsr, then identifies the core chemical structure with \core, finally synthesizing the information into a unified response. More intuitive workflow can be viewed in Figure~\ref{fig:workflow}.

\subsection{\qa}
\label{sec:qa}


\qa is designed to interpret and respond to natural language queries about patents, extracting precise information from the text of patent documents. After collecting user's request from LLM Orchestrator, \qa focuses on specific sections of the patent that are most likely to contain relevant answers. When retrieving relevant sections, \qa utilizes a doc-layout model comprised of object detection and Optical Character Recognition (OCR) to extract specific regions in the patents, such as the abstract, claims, and detailed description.

After getting specific section, \qa employs a transformer-based retrieval model that performs semantic search across the parsed sections of the patent document.  Once the relevant information are retrieved, \qa synthesizes into a coherent answer. If necessary, the LLM orchestra would rephrase the text to ensure the response is directly aligned with the user's query.

\subsection{\ocsr}
\label{sec:ocsr}

The task of \ocsr is to convert images to molecular sequences, which is a task of Optical Chemical Structure Recognition (OCSR). Here specifically we wish to generate Simplified Molecular Input Line Entry System (SMILES)~\cite{weininger1988smiles}.

\begin{figure}[ht]
    \centering
    \includegraphics[width=\columnwidth]{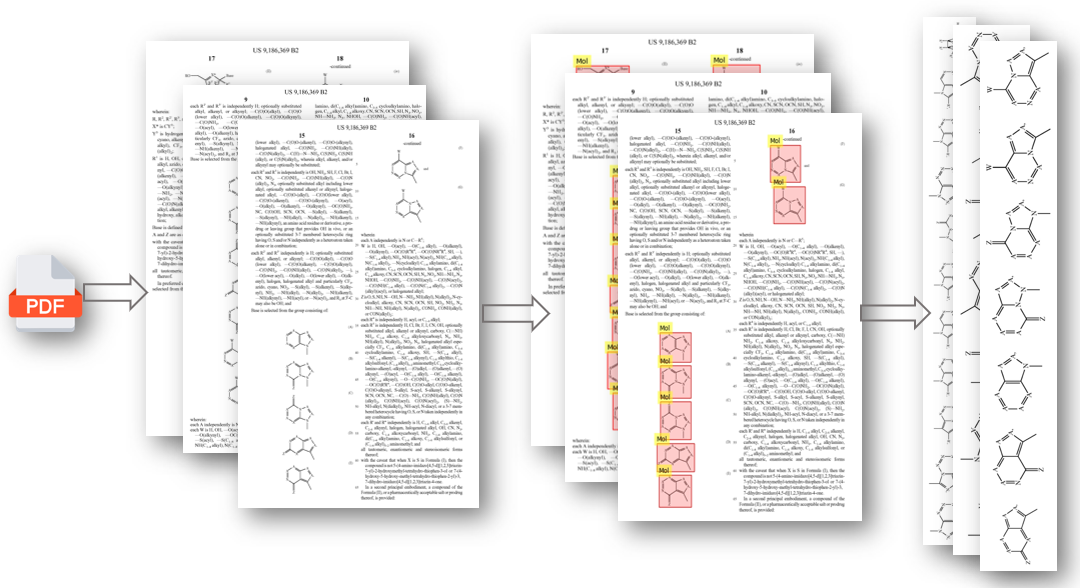}
    \caption{This figure shows the process of extracting related images of chemical structure in pdf patent file.}
    \label{fig:extraction}
\end{figure}

Firstly, We use MolDetect~\cite{MolDetect}\footnote{Github repo forked from RxnScribe~\cite{RxnScribe}} to extract specially images of chemical compounds from patents. An intuitive visual illustration can be viewed in Figure~\ref{fig:extraction}. After retrieving all the chemical images, we followe Algorithm~\ref{alg:ocr} to convert chemical images to SMILES.

\begin{figure*}
    \centering
    \includegraphics[width=\textwidth]{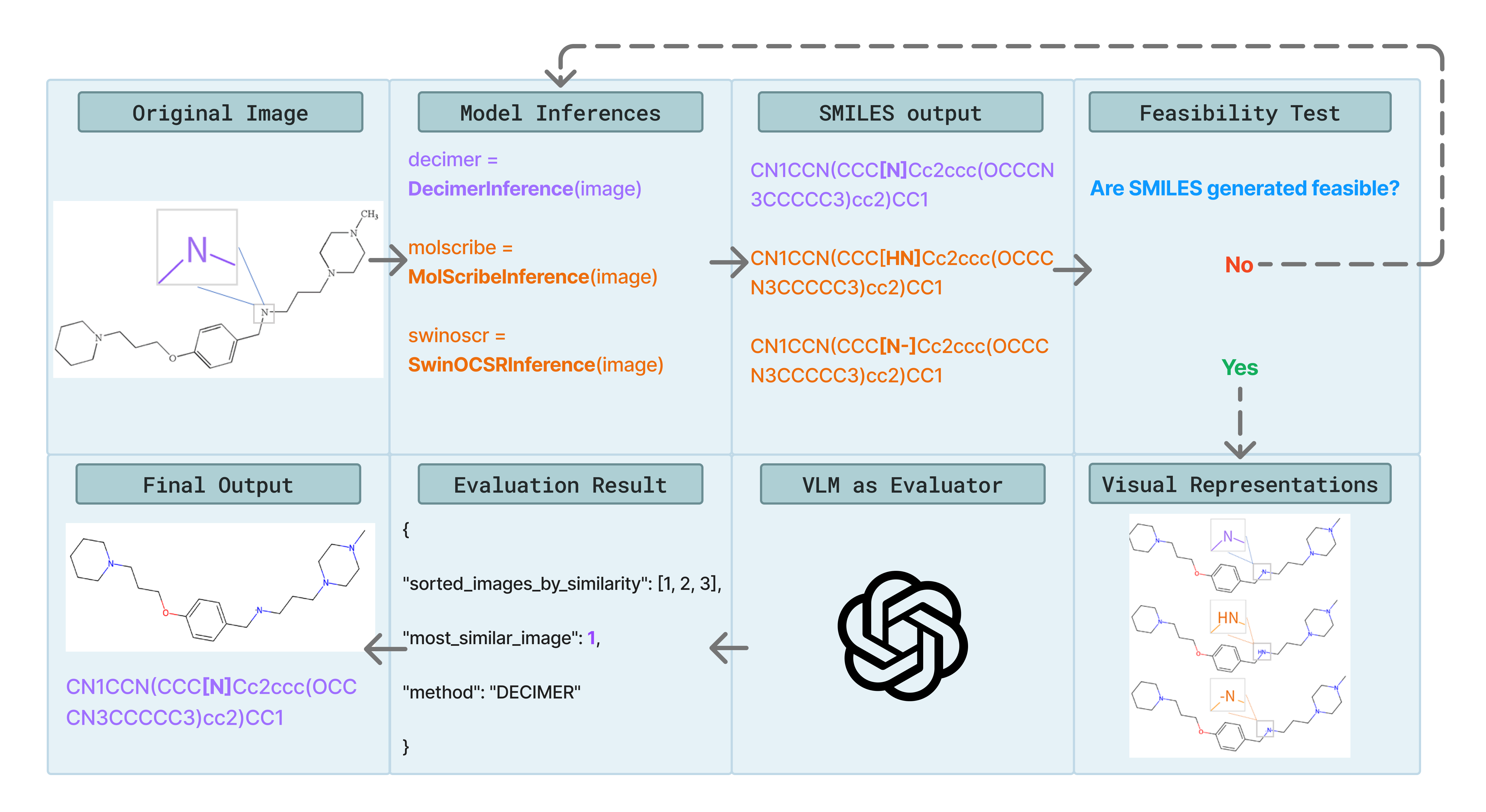}
    \caption{Complete workflow of converting a molecular image to SMILES: from model inference through VLM evaluation to final output.}
    \label{fig:img2mol}
\end{figure*}

Considering the different architectures and training data of each OCSR model, it is natural that each model has its own strengths and weaknesses in learning and identifying the relevant features. Therefore, we select three OCSR models, namely DECIMER 2.0~\cite{rajan2021decimer}, MolScribe~\cite{qian2023molscribe}, and SwinOCSR~\cite{xu2022swinocsr} for the best performance and use GPT-4o~\cite{achiam2023gpt} as the VLM evaluator. An intuitive illustration of the framework can be viewed in Figure~\ref{fig:img2mol}.

\begin{algorithm}[t]
\SetAlgoLined
Given three Image-to-SMILES models $M_1$, $M_2$, and $M_3$ and an input image $I$, this algorithm outputs the final SMILES: $\text{SMILES}_{\text{final}}$\\
\For{each model $M$}{
    $\text{SMILES}_M$ = $f_M(I)$ // \text{Image to SMILES}\\
    $\text{SMILES}_{\text{std}}$ = $g(\text{SMILES}_M)$ // \text{Standardizes}\\
    $I'= \text{RDKit}(\text{SMILES)}$ // \text{Convert back to image}\\
    $\text{similarity}_{M}$ = $\text{VLM}(I, I')$\\
}
$\text{SMILES}_{\text{final}}$ = $\text{SMILES}_{M*}$, when $\text{similarity}_{M*}$ = $\text{max}\{\text{similarity}_{M_1}, \text{similarity}_{M_2}, \text{similarity}_{M_3}\}$
\caption{The OCSR Algorithm with Model Ensemble and VLM as Evaluator}
\label{alg:ocr}
\end{algorithm}

In this module, for each chemical image, we use the three models: DECIMER, MolScribe, and SwinOCSR, to generate SMILES. After standardizing the generated SMILES, we use RDKit~\cite{rdkit} to convert SMILES back to chemical images. Then we use GPT-4o~\cite{achiam2023gpt} the VLM to calculate the similarity between the regenerated image and the original image, so that the similarity considers possible re-orientation and scaling. By comparing the results, we are able to find the best SMILES with the highest similarity.

\subsection{\core}
\label{sec:core}

\core is designed to identify the core chemical structure when given a list of chemicals, most possibly retrieved from the same patent file. The mechanism for identifying the core structure is to find the common structure in all the chemicals, with the understanding that the maximal common substructure is highly likely the core structure~\cite{zhu2024patentnetml}.

\begin{figure*}[th]
    \centering
    \includegraphics[width=\textwidth]{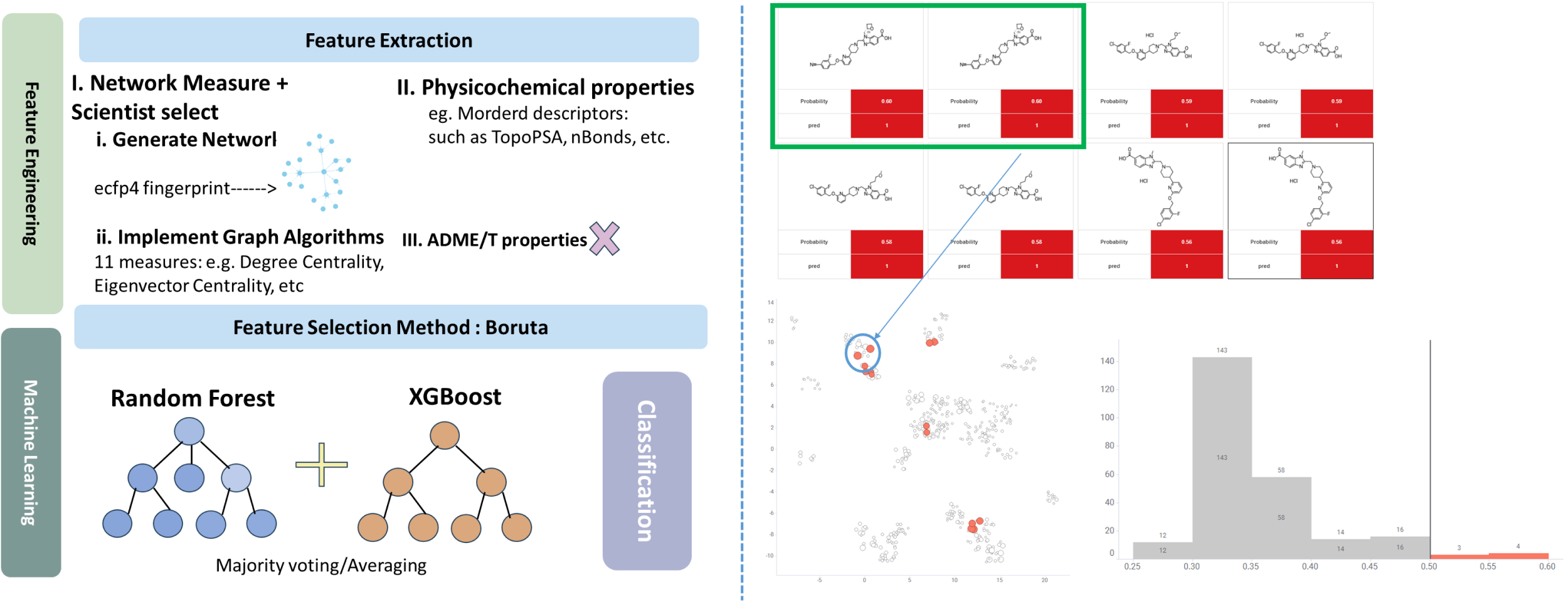}
    \caption{The left image illustrates the framework for achieving specific lead identification (leadid), covering the process from feature selection to model training. The right image shows an example of leadid classification within a patent: (a) in the top left visualizes the lead id, (b) in the bottom left projects the molecules from the patent and marks the position of the lead id and (d) in the top right displays a probability distribution graph of the molecules in the patent.}
    \label{fig:enter-label}
\end{figure*}

To identify the core compound, we trained a machine learning model which ensembles XGBoost~\cite{Chen2016XGBoostAS} and Random Forest~\cite{breiman2001random}. Bayesian Optimization was used to identify the optimal parameters for each model. Boruta~\cite{kursa2010boruta} is used for feature selection. We trained the model with mini dataset\footnote{The mini train consists of only 112 data points} of PatentNetML~\cite{zhu2024patentnetml} with 3080ti for one day.

In preparing the features for the model, considering the focus of our application in pharmaceutical patents, we selected Extended-Connectivity FingerPrints 4 (ECFP4)~\cite{rogers2010extended}, adapted from Morgan algorithm~\cite{morgan1965generation}, to capture the structural information of the chemicals, as ECFP4 is the algorithm that captures the most information. Then we convert the ECFP4 fingerprint and physicochemical properties into a graph network, using cutoff values from 0.4 to 0.9. We also create an adjacency matrix where each pair of vertices is labeled 1 or 0 depending on whether they are adjacent.

\section{Experiments}
\label{sec:exp}

We design experiments to evaluate \mname to answer the following research questions:
\begin{enumerate}[leftmargin=10mm, nosep]
    \item[\textbf{RQ1}:] How effectively does \ocsr convert patent images to molecular structures compared to existing methods?
    \item[\textbf{RQ2}:] How accurately does \core identify core chemical structures within pharmaceutical patents, comparing to other state-of-the-art methods?
    \item[\textbf{RQ3}:] How does \mname perform in a real-world application, such as analyzing a complex pharmaceutical patent, and what insights can it provide to researchers?
\end{enumerate}

\subsection{RQ1: \ocsr converts patent images to molecular structure more effectively than existing methods} 


\noindent
To answer \textbf{RQ1}, we compared \mname with other openly available tools for Optical Chemical Structure Recognition (OCSR):
\begin{enumerate}[leftmargin=*,noitemsep, topsep=0pt]
    \item \textbf{OSRA}:~\citet{filippov2009optical} propose the first open-source tool to identify chemical structure from images using traditional computational techniques.
    \item \textbf{Imago}: \citet{smolov2011imago} suggest a novel algorithm for processing images of chemicals by separating out the graphical and symbolic layers.
    \item \textbf{MolVec}: National Center for Advancing Translational Sciences (\citet{molvec}) publishes a public repository vectorizing chemical images into chemical objects.
    \item \textbf{Img2Mol}: \citet{clevert2021img2mol} train and publicize a deep convolutional neural network that consists of an encoder that converts images into latent space and a decoder that translates latnent space information into SMILES.
    \item \textbf{SwinOCSR}: \citet{xu2022swinocsr} propose a Swin transformer based approach to train an end-to-end model.
    \item \textbf{MolScribe}: \citet{qian2023molscribe} present a generational model that predicts atom, bond, and layer to construct molecular structure.
    \item \textbf{DECIMER}: \citet{rajan2021decimer} brings an image transformer that trains on PubChem~\cite{kim2023pubchem} dataset and a dataset generated with Markush structure.
\end{enumerate}
Experiments comparing \mname and the above models run on these datasets with real chemical structure images without distortion \footnote{These datasets are all obtained from validation datasets at OCSR Benchmark~\cite{Rajan2020}}:
\begin{enumerate}[leftmargin=*,noitemsep, topsep=0pt]
    \item \textbf{CLEF}: 992 images and chemical molecule file pairs published by the Conference and Labs of the Evaluation Forum (CLEF) in 2012.  
    \item \textbf{JPO}: 450 images and chemical molecule file pairs published by Japanese Patent Office (JPO).
    \item \textbf{UOB}: 5470 image and chemical molecule file pairs published by the University of Birmingham (UOB), United Kingdom.
    \item \textbf{USPTO}: 5719 image and chemical molecule file pairs published by the US PaTent Office (USPTO).
    \item \textbf{ACS}: 331 image and chemical molecule file pairs published by American Chemistry Society (ACS) and manually annotated by MolScribe~\cite{qian2023molscribe}.
\end{enumerate}

\begin{table}[]
\resizebox{\columnwidth}{!}{%
\begin{tabular}{lccccc} 
\toprule
          & \textbf{CLEF} & \textbf{JPO} & \textbf{UOB} & \textbf{USPTO}  & \textbf{ACS} \\ \hline
OSRA      & 84.6         & 55.3        & 78.5         & 87.4             & 55.3    \\
MolVec    & 82.8         & 67.8        & 80.6         & \textbf{88.4}    & 47.4    \\
Imago     & 59.0         & 40.0        & 58.0         & 87.0             & -     \\
Img2Mol   & 18.3         & 16.4        & 68.7         & 26.3             & 23.0  \\
SwinOCSR  & 30.0         & 13.8        & 44.9         & 27.9             & 27.5  \\
MolScribe & 88.9         & 76.2        & 87.9         & 79.0             & 71.9    \\
DECIMER   & 62.7         & 55.2        & 88.2         & 41.1             & 46.5  \\
\mname    & \textbf{89.28}& \textbf{78.68}  & \textbf{96.57}    & 82.63     &  \textbf{72.34}\\
\bottomrule
\end{tabular}%
}
\caption{This table presents the experiment results of converting chemical structure images to SMILES. \ocsr recognizes optical chemical structures more accurately than all the other strong baseline models on four of the five benchmarks and competitively on USPTO. }
\label{tab:OSCR-Results}
\end{table}

The results of the experiments are presented in Table~\ref{tab:OSCR-Results}. The results show that \ocsr achieves better performance in four of the five benchmarks. The lower performances of all models the other models on JPO are likely due to the noises in the images, including unsegmented labels, numbers, and English or Japanese characters~\cite{rajan2023decimer, Rajan2020}. The same happens in our experiments too. When using only \ocsr module, the accuracy for JPO benchmark is only 58\%. However, when running the benchmark using the whole agent, the accuracy boosts to 78.68\%, which is the highest of existing tools, which further validates the effectiveness of our \mname framework. 

Moreover, for the USPTO benchmark, following the remarks of~\citet{clevert2021img2mol, rajan2023decimer} all three models surpassing us, OSRA, MolVec, and Imago, are overfitted to available benchmarks. While our \ocsr contains no training process and utilizes none of the overfitted models, it still reaches competitive accuracy on USPTO benchmark.

\subsection{RQ2: \core identifies core chemical structure more accurately comparing to other SoTA methods}

To answer \textbf{RQ2}, we compared our model with PatentNetML~\cite{zhu2024patentnetml} and other traditional Cheminformatics Methods, including Cluster Seed Analysis(CSA)~\cite{hattori2008predicting}, Molecular Idol(MI)~\cite{lee2019idol}, and Frequency of Group Analysis(FOG)~\cite{tyrchan2012exploiting} on PatentNetML dataset~\cite{zhu2024patentnetml}.

\begin{table}[]
\resizebox{\columnwidth}{!}{%
\begin{tabular}{lccc|cc}
\toprule
(Top)        & \textbf{1}    & \textbf{5}     & \textbf{10}     & \textbf{5\%}     & \textbf{10\%}\\ \hline
CSA         & 6.14               & 27.00           & -                &   -               &   -             \\
MI          & 5.41               & 18.00           &  -               & -                 & -               \\
FOG         & 6.37               & 26.00           &  -               & -                 &    -            \\
PatentNetML & 7.14         & 35.71        & 50.00         & 28.57          & 50.00        \\
\mname      & 14.29        & 35.71        & 50.00         & 35.71          & 50.00        \\
\bottomrule
\end{tabular}%
}
\caption{\core predicts the core chemical in patents more accurately than the recent machine-learning and cheminformatics approaches. It reaches SoTA in all accuracy measures.}
\label{tab:Core-Results}
\end{table}

We present the results from the experiments in Table~\ref{tab:Core-Results} with all the accuracy measures, including Top 1, Top 5, Top 10, Top 1\%, Top 5\%, and Top 10\%. We got the highest percentage in all these accuracy measures with the updated feature selection algorithm.

\subsection{RQ3: \mname improves the efficiency when dealing with complex patent --- a case study}
\label{sec:case}

We select the patent with ID US-8063224~\cite{lachance_azacycloalkane_2011} for this case study. This patent covers azacycloalkane derivatives acting as selective inhibitors intended for the treatment and prevention of diseases related to abnormal lipid metabolism, such as cardiovascular disease, obesity, diabetes, and metabolic syndrome.

To demonstrate \mname's ability under real-world circumstances, we want to test if \mname successfully complete the requests of the users and achieves high accuracy when identifying core chemical structure.

When asked to identify the core chemical compound from this patent, after scanning the file, extracting chemical compound images, converting these images to SMILES, and identifying the top ten candidates with the highest probabilities, the core chemical compound ranks sixth. This lower ranking compared to using \core alone is due to discrepancies in retrieving all chemical images, as many chemicals in the patent are not listed as images. Nevertheless, \mname effectively reduces the effort needed to identify the core chemical compound from fifty compounds, saving time for researchers in testing and linking tools.

\begin{figure*}[ht]
    \centering
    \includegraphics[width=\textwidth]{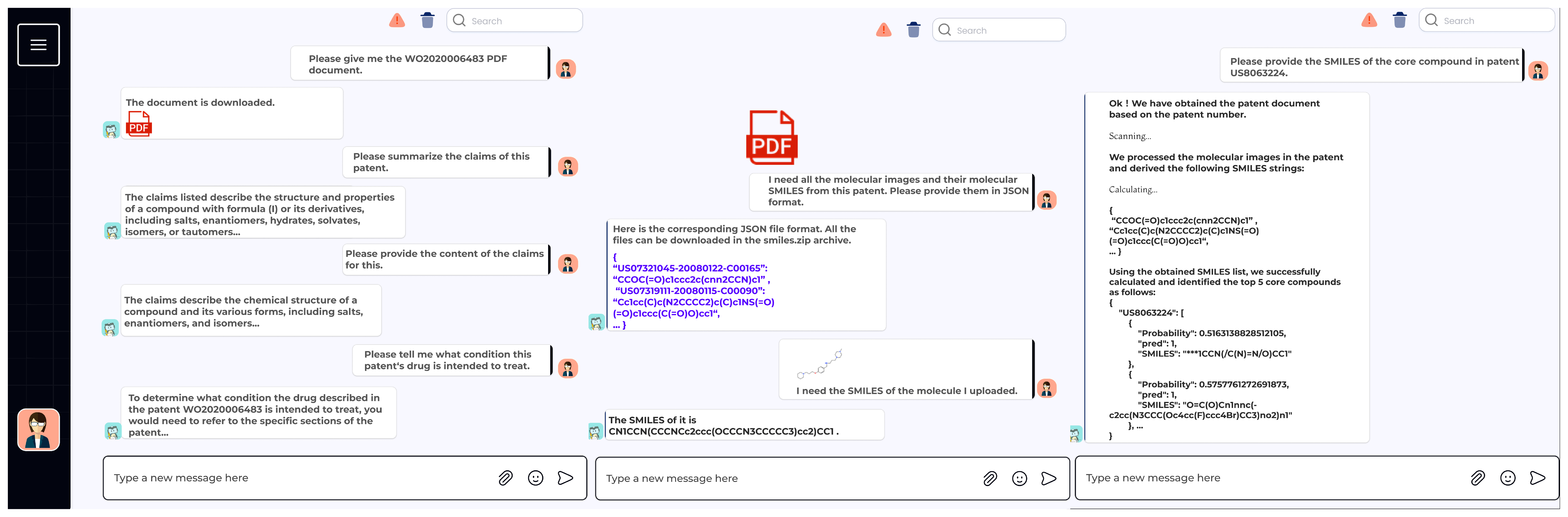}
    \caption{User Interface of \mname.}
    \label{fig:case-study}
\end{figure*}

\section{Discussion}
\label{sec:discussion}

Our experiments show that \mname is a robust tool capable of addressing key aspects of patent analysis within a single, integrated framework. The effective operation across core modules—patent question answering, image-to-molecular-structure conversion, and chemical identification—demonstrates its functionality and ability to streamline traditionally fragmented workflows.

\textbf{Integration of Multiple Aspects.} One of the most significant finding is \mname's ability to unify diverse tasks into a unified system, effectively bridging gaps that usually require multiple tools. This integration ensures consistency and enhances efficiency in patent analysis, suggesting that \mname could serve as an all-in-one solution.

\textbf{Accuracy Boosts.} Our experiments, particularly those conducted on the JPO dataset, indicate that \mname improves accuracy in identifying and analyzing chemical structures within patents, leading to more reliable insights for subsequent research. 

\textbf{Efficiency in Real-Life Applications.} \mname simplifies tasks traditionally handled by separate modules into one streamlined step, significantly reducing time and effort. This makes patent analysis more accessible and manageable, broadening its use among researchers.

These findings collectively reinforce the initial claim that \mname has the potential to transform the drug discovery routine by accelerating the process from patent analysis to actionable insights, contributing to faster and more effective drug development.

\section{Related Works}
\label{sec:related}

This section provides an overview of existing approaches in patent analysis in time and method development order, along with gaps that our approach aims to address.

\subsection{Text Mining and Natural Language Processing}

With advancements in computational linguistics, NLP techniques have been applied to patent analysis. SCRIPDB provides a database that automatically retrieves relevant patent information, including syntheses, chemicals, and reactions \cite{heifets2012scripdb}. Chemical NER methods like tmChem \cite{leaman2015tmchem} and ChemSpot \cite{rocktaschel2012chemspot} efficiently identify chemical names in patent text. \citet{feng2012patent} combines text mining with informetric methods for morphological analysis of patent technology. \citet{chikkamath2020empirical} introduces novelty detection and evaluates various language models. \citet{kronemeyer2021stimulating} presents a process model identifying frugal patents.

Recent works leverage AI and language models for patent innovation and validation. Inspired by TRIZ \cite{ilevbare2013review}, \citet{Trapp2024LLMbasedEO} identifies contradictions in patents using LLMs to drive technological progress. PARIS and LE-PARIS generate effective responses to Office Actions (OAs) \cite{Chu2024FromPT}, while PRO combines legal knowledge graphs with LLMs through retrieval-augmented generation (RAG) for more faithful OA responses \cite{Chu2024PatentRS}.

\subsection{Chemical Structure Analysis}
Chemical structures are the vital treasure of information in biochemical patents, especially in drug discoveries. There are plenty of works related to chemical structures in patents, and the first one is mining chemicals in patents.

SureChEMBLcss~\cite{falaguera2021identification} contains authentic information on core chemicals in patents in the SureChEMBL datasets. PatentNetML~\cite{zhu2024patentnetml} predicts key compound using machine learning networks, a model MOMP~\cite{turutov2024molecular} is proposed to optimize molecules with patentability constraint. CASTER~\cite{huang2020caster} predicts drug-drug interactions (DDIs) given chemical structures of drugs.

\subsection{Integrated AI Systems}
The increasing complexity and volume of patent data, with the development of large language models (LLMs), have led to increasing efforts in integrating AI into a whole system consisting of multiple analytical modules.
MolTailor~\cite{guo2024moltailor} combines molecular representation model as a knowledge base with language model as an agent to achieve molecule-text multi-task regression (MT-MTR). DECIMER.ai~\cite{rajan2023decimer} is an open platform for automatic chemical structure segmentation, classification, and translation in scientific publications, which can also be generalized into patents.

\section{Conclusion}
\label{sec:conclusion}

In conclusion, in this paper we propose \mname as a novel approach in utilizing Large Language Models (LLMs) to link vital tools in pharmaceutical patent analysis so that researchers and practitioners can interactively and efficiently analyze patents from both text and chemical compounds. Moreover, \ocsr and \core with updated algorithm achieve better performances than current tools. The case study also presents as a qualitative validation of the real-life application of \mname even under unusual and complex circumstances.

While our results are encouraging, they represent just the beginning of what \mname could achieve. With further development and refinement, \mname has the potential to not only streamline patent analysis but also fundamentally change the drug discovery routine by making the process faster, more accurate, and more accessible to researchers worldwide. As the field of automatic patent analysis develops, we wish to see more evaluation datasets and benchmarks for assessing intelligent automatic patent analysis agent like \mname.

\appendix


\bibliography{references}

\end{document}